\documentclass[11pt,a4paper]{article}
\pdfoutput=1

\usepackage[hyperref]{naaclhlt2018}
\usepackage{times}
\usepackage{latexsym}

\usepackage{url}
\usepackage{amsmath}
\usepackage{multirow}
\usepackage{booktabs}
\usepackage[T1]{fontenc}
\usepackage{microtype}

\usepackage{subcaption}
\usepackage{filecontents}

\aclfinalcopy 
\def\aclpaperid{980} 

\usepackage{xspace}
\newcommand{\mtwo}{M$^2$\xspace}

\newcommand{\err}[1]{\textit{#1}}
\newcommand{\errcor}[2]{\textit{#1}$\rightarrow$\textit{#2}}

\usepackage{graphicx}
\usepackage{tikz}
\usepackage{pgfplots}
\usepackage{pgfplotstable}
\pgfplotsset{compat=1.12}
\usepgfplotslibrary{groupplots}
\pgfplotsset{try min ticks=6}

\title{Near Human-Level Performance in Grammatical Error Correction\\ with Hybrid Machine Translation}

\author{Roman Grundkiewicz \\
        University of Edinburgh \\
        10 Crichton St, Edinburgh EH8 9AB, Scotland \\
        {\tt rgrundki@inf.ed.ac.uk} \\\And
  Marcin Junczys-Dowmunt \\
  Microsoft \\
  Redmond, WA 98052, USA \\
  {\tt marcinjd@microsoft.com} \\}

\date{}

\begin{document}
\maketitle
\begin{abstract}
We combine two of the most popular approaches to automated Grammatical Error Correction (GEC): GEC based on Statistical Machine Translation (SMT) and GEC based on Neural Machine Translation (NMT). The hybrid system achieves new state-of-the-art results on the CoNLL-2014 and JFLEG benchmarks. This GEC system preserves the accuracy of SMT output and, at the same time, generates more fluent sentences as it typical for NMT.
Our analysis shows that the created systems are closer to reaching human-level performance than any other GEC system reported so far.
\end{abstract}

\section{Introduction}

\begin{figure}[t]\centering\small%
  \begin{tikzpicture}
  \begin{axis}[
    title={SMT $\quad\quad\quad\;\;$ NMT $\quad\quad\quad$ Hybrid},
    enlarge y limits,
    ymajorgrids,
    major grid style={dotted},
    y tick label style={/pgf/number format/.cd, fixed, fixed zerofill, precision=0, /tikz/.cd},
    ylabel=M\textsuperscript{2},
    every axis y label/.style={ at={(ticklabel* cs:.95)}, anchor=south east},
    xtick={1,2,3, 4.8,5.6,6.4,7.2, 9,10,11},
    xticklabels={
        R\&R'16,JD\&G'16,This work,
        Y\&B'16,Sch.\&al.'17,Ji\&al.'17,This work,
        Y\&al.'17,Ch.\&Ng'17,This work
    },
    xmin= 0.2,
    xmax= 11.8,
    ymin=38,
    ymax=57,
    width=.48\textwidth,
    height=0.25\textheight,
    xticklabel style={align=right, rotate=45, anchor=north east},
    legend pos= north west,
    legend style={cells={anchor=west}, at={(0.02,0.98)}}
      ]
    \addplot+[black,only marks,mark options={fill=white},solid,mark=*]%
    table[row sep=\\,x=x,y=y] { x     y\\ 1  47.40\\};
    \addplot+[black,only marks,mark options={fill=white},solid,mark=*]
    table[row sep=\\,x=x,y=y] { x     y\\ 2  49.49\\ 2 48.49\\};
    \addplot+[black,only marks,solid,mark=*,mark options={fill=black}] 
    table[row sep=\\,x=x,y=y] { x     y\\ 3  50.27\\ 3 49.82\\ 3 48.38\\};

    \addplot+[mark=none,dashed]
    table[row sep=\\,x=x,y=y] {x      y\\ 4  0\\ 4 100\\};

    \addplot+[black,only marks,mark options={fill=white},solid,mark=*]
    table[row sep=\\,x=x,y=y] { x     y\\ 4.8  40.56\\};
    \addplot+[black,only marks,mark options={fill=white},solid,mark=*]
    table[row sep=\\,x=x,y=y] { x     y\\ 5.6  41.37\\};
    \addplot+[black,only marks,mark options={fill=white},solid,mark=*]
    table[row sep=\\,x=x,y=y] { x     y\\ 6.4  45.15\\ 6.4 44.99\\ 6.4 42.82\\};
    \addplot+[black,only marks,solid,mark=*,mark options={fill=black}] 
    table[row sep=\\,x=x,y=y] {x      y\\ 7.2  50.19\\ 7.2 48.56\\ 7.2 42.76\\ 7.2 41.29\\};

    \addplot+[black,mark=none,dashed]
    table[row sep=\\,x=x,y=y] {x      y\\ 8  0\\ 8 100\\};

    \addplot+[black,only marks,mark options={fill=white},solid,mark=*]
    table[row sep=\\,x=x,y=y] { x     y\\ 9  51.08\\ 9 49.66\\};
    \addplot+[black,only marks,mark options={fill=white},solid,mark=*]
    table[row sep=\\,x=x,y=y] { x     y\\ 10  53.14\\ 10 51.70\\};
    \addplot+[black,only marks,solid,mark=*,mark options={fill=black}] 
    table[row sep=\\,x=x,y=y] {x      y\\ 11  56.25\\ 11 56.05\\ 11 54.95\\ 11 53.51\\};

  \end{axis}
  \end{tikzpicture}
  \caption{Comparison of SMT, NMT and hybrid GEC systems on the CoNLL-2014 test set (\mtwo).}
  \label{tab:relatedwork}
\end{figure}
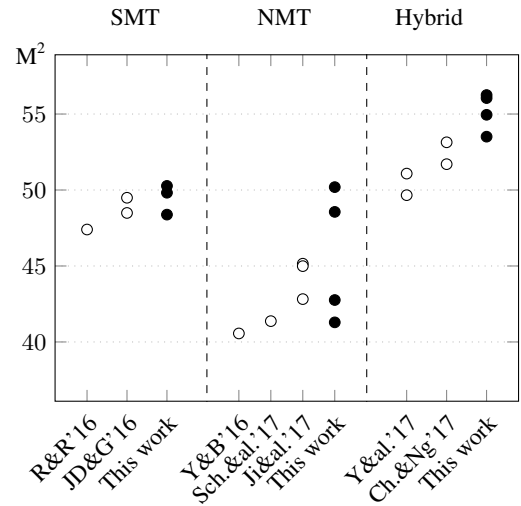

Currently, the most effective GEC systems are based on phrase-based statistical machine translation \cite{rozovskaya2016grammatical,junczys2016phrase,chollampatt2017connecting}.
Systems that rely on neural machine translation \cite{yuan2016grammatical,xie2016neural,schmaltz2017adapting,ji2017nested} are not yet able to achieve as high performance as SMT systems according to automatic evaluation metrics (see Table~\ref{tab:relatedwork} for comparison on the CoNLL-2014 test set).
However, it has been shown that the neural approach can produce more fluent output, which might be desirable by human evaluators \cite{napoles2017jfleg}.
In this work, we combine both MT flavors within a hybrid GEC system.
Such a GEC system preserves the accuracy of SMT output and at the same time generates more fluent sentences achieving new state-of-the-art results on two different benchmarks: the annotation-based CoNLL-2014 and the fluency-based JFLEG benchmark.
Moreover, comparison with human gold standards shows that the created systems are closer to reaching human-level performance than any other GEC system described in the literature so far.

Using consistent training data and preprocessing (\autoref{sec:data}), we first create strong SMT (\autoref{sec:smt}) and NMT (\autoref{sec:nmt}) baseline systems.
Then, we experiment with system combinations through pipelining and reranking (\autoref{sec:hybrid}).
Finally, we compare the performance with human annotations and identify issues with current state-of-the-art systems (\autoref{sec:analysis}).

\section{Data and preprocessing}
\label{sec:data}

Our main training data is NUCLE \cite{dahlmeier2013building}.
English sentences from the publicly available Lang-8 Corpora \cite{mizumoto2012effect} serve as additional training data.

We use official test sets from two CoNLL shared tasks from 2013 and 2014 \cite{ng2013conll,ng2014conll} as development and test data, and evaluate using \mtwo \cite{dahlmeier2012better}.
We also report results on JFLEG \cite{napoles2017jfleg} with the GLEU metric \cite{napoles2015ground}.
The data set is provided with a development and test set split.
All data sets are listed in Table~\ref{tab:data}.

We preprocess Lang-8 with the NLTK tokenizer \cite{bird2004nltk} and preserve the original tokenization in NUCLE and JFLEG.
Sentences are truecased with scripts from Moses \cite{koehn2007moses}.
For dealing with out-of-vocabulary words, we split tokens into 50k subword units using Byte Pair Encoding (BPE) by \newcite{sennrich2016bpe}. BPE codes are extracted only from correct sentences from Lang-8 and NUCLE.

\begin{table}[t]\centering
\small
\begin{tabular}{lrrr}
\toprule
    Corpus             & Sentences &     Tokens \\
\midrule
    NUCLE              &    57,151 &  1,162K \\
    Lang-8 NAIST       & 1,943,901 & 25,026K \\
    CoNLL-2013 (dev)   &     1,381 &     29K \\
    CoNLL-2014 (test)  &     1,312 &     30K \\
    JFLEG Dev          &       754 &     14K \\
    JFLEG Test         &       747 &     13K \\
\bottomrule
\end{tabular}
    \caption{Statistics for training and testing data sets.}
\label{tab:data}
\end{table}

\section{SMT systems}
\label{sec:smt}

For our SMT-based systems, we follow recipes proposed by \newcite{junczys2016phrase}, and use a phrase-based SMT system with a log-linear combination of task-specific features.
We use word-level Levenshtein distance and edit operation counts as dense features (Dense), and correction patterns on words with one word left/right context on Word Classes (WC) as sparse features (Sparse).
We also experiment with additional character-level dense features (Char. ops).
All systems use a 5-gram Language Model (LM) and OSM \cite{durrani2011joint} both estimated from the target side of the training data, and a 5-gram LM and 9-gram WCLM trained on Common Crawl data \cite{buck2014ngram}.

\paragraph{Experiment settings}

Translation models are trained with Moses \cite{koehn2007moses}, word-alignment models are produced with MGIZA++ \cite{gao2008parallel}, and no reordering models are used.
Language models are built using KenLM \cite{heafield2011kenlm}, while word classes are trained with word2vec\footnote{\url{https://github.com/dav/word2vec}}.

We tune the systems separately for \mtwo and GLEU metrics.
MERT \cite{och2003minimum} is used for tuning dense features and Batch Mira \cite{cherry2012batch} for sparse features.
For \mtwo tunning we follow the 4-fold cross-validation on NUCLE with adapted error rate recommended by \newcite{junczys2016phrase}.
Models evaluated on GLEU are optimized on JFLEG Dev using the GLEU scorer, which we added to Moses. We report results for models using feature weights averaged over 4 tuning runs.

\begin{table}[t]\centering
\small
    \begin{tabular}{lp{0.5cm}p{0.5cm}p{0.5cm}c}
\toprule
           & \multicolumn{3}{c}{CoNLL} & \multicolumn{1}{c}{JFLEG} \\
    System & \multicolumn{1}{c}{P} & \multicolumn{1}{c}{R} & \multicolumn{1}{c}{\mtwo} & \multicolumn{1}{c}{GLEU} \\
\midrule
    SMT Dense           & 56.91 & 30.25 & 48.38      & 54.68      \\
    \quad + Sparse      & 60.28 & 29.40 & \bf{49.82} & 55.25      \\
    \quad\quad + Char. ops   & 60.27 & 30.21 & \bf{50.27} & \bf{55.79} \\
\bottomrule
\end{tabular}
    \caption{Results for SMT baseline systems on the CoNLL-2014 (\mtwo) and JFLEG Test (GLEU) sets.}
\label{tab:smt}
\end{table}

\paragraph{Results}

Other things being equal, using the original tokenization, applying subword units, and extending edit-based features result in a similar system to \newcite{junczys2016phrase}: 49.82 vs 49.49 \mtwo (Table~\ref{tab:smt}).

The phrase-based SMT systems do not deal well with orthographic errors \cite{napoles2017jfleg} --- if a source word has not been seen in the training corpus, it is likely copied as a target word.
Subword units can help to solve this problem partially.
Adding features based on character-level edit counts increases the results on both test sets. 

A result of 55.79 GLEU on JFLEG Test is already 2 points better than the GLEU-tuned NMT system of \newcite{sakaguchi-post-vandurme:2017:I17-2} and only 1 point worse than the best reported result by \newcite{chollampatt2017connecting} with their \mtwo-tuned SMT system, even though no additional spelling correction has been used at this point. We experiment with specialized spell-checking methods in later sections.

\section{NMT systems}
\label{sec:nmt}

The model architecture we choose for our NMT-based systems is an attentional encoder-decoder model with a bidirectional single-layer encoder and decoder, both using GRUs as their RNN variants \cite{sennrich2017nematus}.
A similar architecture has been already tested for the GEC task by \newcite{sakaguchi-post-vandurme:2017:I17-2}, but we use different hyperparameters.

To improve the performance of our NMT models, similarly to \newcite{xie2016neural} and \newcite{ji2017nested}, we combine them with an additional large-scale language model.
In contrast to previous studies, which use an $n$-gram probabilistic LM, we build a 2-layer Recurrent Neural Network Language Model (RNN LM) with GRU cells which we train again on English Common Crawl data \cite{buck2014ngram}.

\paragraph{Experimental settings}
We train with the Marian toolkit \cite{mariannmt} on the same data we used for the SMT baselines, i.e. NUCLE and Lang-8.
The RNN hidden state size is set to 1024, embedding size to 512. Source and target vocabularies as well as subword units are the same.

Optimization is performed with Adam \cite{kingma2014adam} and the mini-batch size fitted into 4GB of GPU memory. We regularize the model with scaling dropout \cite{gal2016theoretically} with a dropout probability of 0.2 on all RNN inputs and states. Apart from that we dropout entire source and target words with probabilities of 0.2 and 0.1 respectively.
We use early stopping with a patience of 10 based on the cross-entropy cost on the CoNLL-2013 test set.
Models are validated and saved every 10,000 mini-batches. As final models we choose the one with the best performance on the development set among the last ten model check-points based on the \mtwo or GLEU metrics.

Size of RNN hidden state and embeddings, target vocabulary, and optimization options for the RNN LM are identical to those used for our NMT models. 
Decoding is done by beam search with a beam size of 12.
We normalize scores for each hypothesis by sentence length.

\begin{table}[t]\centering
  \small
  \begin{tabular}{lp{0.5cm}p{0.5cm}p{0.5cm}c}
    \toprule
             & \multicolumn{3}{c}{CoNLL} & \multicolumn{1}{c}{JFLEG} \\
      System & \multicolumn{1}{c}{P} & \multicolumn{1}{c}{R} & \multicolumn{1}{c}{\mtwo} & \multicolumn{1}{c}{GLEU} \\
    \midrule
    NMT                               & 66.61 & 17.58 & 42.76         & 50.08       \\
    NMT + RNN-LM                      & 61.05 & \bf{26.71} & 48.56    & 56.04       \\ 
    NMT$\times$4                      & \bf{71.10} & 15.42 & 41.29    & 50.30       \\
    NMT$\times$4 + RNN-LM             & 60.27 & 30.08 & \bf{50.19}    & \bf{56.74}  \\ 
  \bottomrule
  \end{tabular}
    \caption{Results for NMT systems on the CoNLL-2014 (\mtwo) and JFLEG Test (GLEU) sets.}
  \label{tab:nmt}
\end{table}

\paragraph{Results}

A single NMT model achieves lower performance than the SMT baselines (Table~\ref{tab:nmt}).
However, the \mtwo score of 42.76 for CoNLL-2014 is already higher than the best published result of 41.53 \mtwo for a strictly neural GEC system of \newcite{ji2017nested} that has not been enhanced by an additional language model.

Our RNN LM is integrated with NMT models through ensemble decoding \cite{sennrich2016edinburgh}.
Similarly to \newcite{ji2017nested}, we choose the weight of the language model using grid search on the development set%
    \footnote{Used weights are 0.2 and 0.25 for \mtwo and GLEU evaluation, respectively.}.
This strongly improves recall, and thus boosts the results significantly on both test sets (+5.8 \mtwo and +5.96 GLEU).

An ensemble of four independently trained models\footnote{Each model is weighted equally during decoding.} (NMT$\times$4), on the other hand, increases precision at the expense of recall, which may even lead to a performance drop. Adding the RNN LM to that ensemble balances this negative effect, resulting in 50.19 \mtwo. These are by far the highest results reported on both benchmarks for pure neural GEC systems.

\paragraph{Comparison to SMT systems}

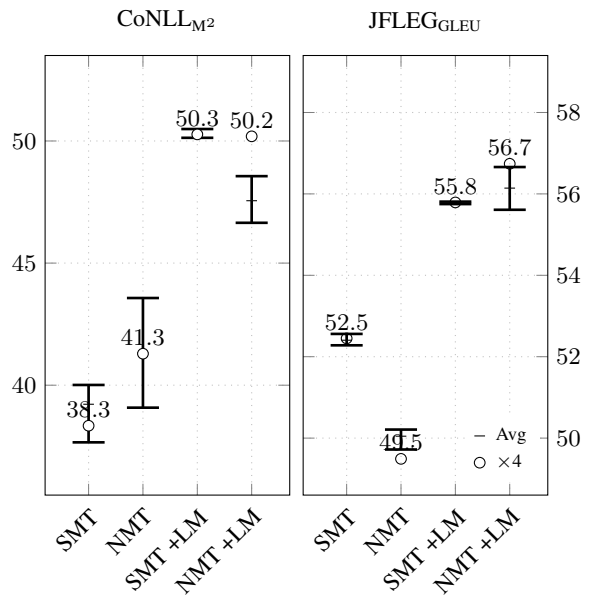
\begin{figure}[t]\centering\small%

\begin{tikzpicture}%
\begin{axis}[
title={CoNLL$_{\text{M}^2}$},
every axis y label/.style={ at={(ticklabel* cs:0.93)}, anchor=south east, align=right },
height=0.3\textheight,
width=0.3\textwidth,
enlarge y limits,
ymajorgrids, xmajorgrids,
major grid style={dotted},
xtick={1,2,3,4},
xticklabels={SMT,NMT,SMT +LM,NMT +LM},
xmin=0.2,
xmax=4.7,
ymin=37,
ymax=52,
xticklabel style={align=right, rotate=45, anchor=north east},
every node near coord/.append style={anchor=south, font=\small, /pgf/number format/.cd, fixed, fixed zerofill, precision=1, /tikz/.cd},
]
\addplot+[black,solid, mark=-, only marks,error bars/.cd,y dir=both,y explicit, error bar style={line width=1pt},
    error mark options={
      rotate=90,
      mark size=6pt,
      line width=1pt
    }]
table[x index=0,
      y expr={(\thisrow{m1}+\thisrow{m2}+\thisrow{m3}+\thisrow{m4})/4},
      y error plus expr={max(\thisrow{m1},\thisrow{m2},\thisrow{m3},\thisrow{m4})-(\thisrow{m1}+\thisrow{m2}+\thisrow{m3}+\thisrow{m4})/4},
      y error minus expr={-min(\thisrow{m1},\thisrow{m2},\thisrow{m3},\thisrow{m4})+(\thisrow{m1}+\thisrow{m2}+\thisrow{m3}+\thisrow{m4})/4},
      ] {compare.test2014};
\addplot+[black,nodes near coords, solid, mark=*, only marks,mark options={fill=white}]
table[x index=0, y=mert] {compare.test2014};
\end{axis}%
\end{tikzpicture}\hfill%
\begin{tikzpicture}%
\begin{axis}[
        title={JFLEG$_{\text{GLEU}}$},
every axis y label/.style={ at={(ticklabel* cs:0.93)}, anchor=south east, align=right, xshift=1.1cm },
cycle list name=black white,
yticklabel pos=right,
height=0.3\textheight,
width=0.3\textwidth,
enlarge y limits,
ymajorgrids, xmajorgrids,
major grid style={dotted},
xtick={1,2,3,4},
xticklabels={SMT,NMT,SMT +LM,NMT +LM},
xmin=0.2,
xmax=4.7,
ymin=49.5,
ymax=58.5,
legend pos=south east,
legend cell align=left,
legend style={column sep=2pt, font=\scriptsize, draw=none},
xticklabel style={align=right, rotate=45, anchor=north east},
every node near coord/.append style={anchor=south, font=\small, /pgf/number format/.cd, fixed, fixed zerofill, precision=1, /tikz/.cd},
]
\addplot+[black,solid, mark=-, only marks,error bars/.cd,y dir=both,y explicit, error bar style={line width=1pt},
    error mark options={ rotate=90, mark size=6pt, line width=1pt }]
table[x index=0,
      y expr={(\thisrow{m1}+\thisrow{m2}+\thisrow{m3}+\thisrow{m4})/4},
      y error plus expr={max(\thisrow{m1},\thisrow{m2},\thisrow{m3},\thisrow{m4})-(\thisrow{m1}+\thisrow{m2}+\thisrow{m3}+\thisrow{m4})/4},
      y error minus expr={-min(\thisrow{m1},\thisrow{m2},\thisrow{m3},\thisrow{m4})+(\thisrow{m1}+\thisrow{m2}+\thisrow{m3}+\thisrow{m4})/4},
      ] {compare.jflegtest};
\addplot+[black,nodes near coords, solid, mark=o, only marks,mark options={fill=white}]
table[x index=0, y=ens] {compare.jflegtest};
\legend{Avg,$\times$4}
\end{axis}%
\end{tikzpicture}
    \caption{Contribution of a language model (LM) for SMT and NMT GEC systems.}
\label{fig:nolm}
\end{figure}

With model ensembling, the neural systems achieve performance similar to SMT baselines (Figure~\ref{fig:nolm}).
A stripped-down SMT system without CCLM, quite surprisingly gives better results on JFLEG than the NMT system, and the opposite is true for CoNLL-2014.
The reason for the lower performance on JFLEG might be a large amount of spelling errors, which are more efficiently corrected by the SMT system using subword units.

If both systems are enhanced by a large-scale language model, the neural system outperforms the SMT system on JFLEG and it is competitive with SMT systems on CoNLL-2014.
However, it is not known if the results would preserve if the NMT model is combined with a probabilistic $n$-gram LM instead as it has been proposed in the previous works \cite{xie2016neural,ji2017nested}.

\section{Hybrid SMT-NMT systems}
\label{sec:hybrid}

We experiment with pipelining and rescoring methods in order to combine our best SMT and NMT GEC systems%
\footnote{The best system combinations are chosen again based on the development sets, i.e.~CoNLL-2013 and JFLEG Dev. We omit these results as they are highly overestimated.}.

\paragraph{SMT-NMT pipelines}

The output corrected by an SMT system is passed as an input to the NMT ensemble with or without RNN LM\footnote{We did not observed any improvements if the order of the systems is reversed.}.
In this case the NMT system serves as an automatic post-editing system.
Pipelining improves the results on both test sets by increasing recall (Table~\ref{tab:hybrid}).
As the performance of the NMT system without a RNN LM is much lower than the performance of the SMT system alone, this implies that both approaches produce complementary corrections.

\begin{table}\centering
\small
\begin{tabular}{lp{0.5cm}p{0.5cm}p{0.5cm}c}
    \toprule
               & \multicolumn{3}{c}{CoNLL} & \multicolumn{1}{c}{JFLEG} \\
        System & P & R & \mtwo & GLEU \\
    \midrule
        Best SMT                                & 60.27 & 30.21 & 50.27         & 55.79 \\
        $\rightarrow$ Pip. NMT                  & 60.25 & 34.80 & 52.56         & 57.21 \\
        $\rightarrow$ Pip. NMT+LM               & 58.87 & \bf{39.23} & 53.51    & \bf{58.83} \\
    \midrule
        + Res. RNN-LM                           & 70.97 & 24.86 & 51.77         & 56.97 \\
        + Res. NMT                              & 70.40 & 26.69 & 53.03         & 57.21 \\
        + Res. NMT+LM                           & 71.40 & \bf{28.60} & \bf{54.95}    & 57.53 \\
    \midrule
        \quad $\rightarrow$ Pip. NMT+LM         & 65.73 & 33.36 & \bf{55.05}    & \bf{58.83} \\
        \quad + Spell SMT                       & 70.80 & 30.57 & 56.05         & 60.09 \\
        \quad\quad $\rightarrow$ Pip. NMT+LM    & 66.77 & 34.49 & \bf{56.25}    & \bf{61.50} \\
  \bottomrule
  \end{tabular}
    \caption{Results for hybrid SMT-NMT systems on the CoNLL-2014 (\mtwo) and JFLEG Test (GLEU) sets.}
\label{tab:hybrid}
\end{table}

\paragraph{Rescoring with NMT}

Rescoring of an n-best list obtained from one system by another is a commonly used technique in GEC, which allows to combine multiple different systems or even different approaches \cite{hoang2016exploiting,yannakoudakis2017neural,chollampatt2017connecting,ji2017nested}.
In our experiments, we generate a 1000 n-best list with the SMT system and add separate scores from each neural component.
Scores of NMT models and the RNN LM are added in the form of probabilities in negative log space.
The re-scored weights are obtained from a single run of the Batch Mira algorithm \cite{cherry2012batch} on the development set.

As opposed to pipelining, rescoring improves precision at the expense of recall and is more effective for the CoNLL data resulting in up to 54.95 \mtwo.
On JFLEG, rescoring only with the RNN LM produces similar results as rescoring with the NMT ensemble.
However, the best result for rescoring is lower than for pipelining on that test set.
It seems the SMT system is not able to produce as diversified corrections in an n-best list as those generated by the NMT ensemble.

\paragraph{Spelling correction and final results}

Pipelining the NMT-rescored SMT system and the NMT system leads to further improvement.
We believe this can be explained by different contributions to precision and recall trade-offs for the two methods, similar to effects observed for the combination of the NMT ensemble and our RNN LM.

On top of our final hybrid system we add a spell-checking component, which is run before pipelining.
We use a character-level SMT system following \newcite{chollampatt2017connecting} which they deploy for unknown words in their word-based SMT system. As our BPE-based SMT does not really suffer from unknown words, we run the spell-checking component on words that would have been segmented by the BPE algorithm.
This last system achieves the best results reported in this paper: 56.25 \mtwo on CoNLL-2014 and 61.50 GLEU on JFLEG Test.

\begin{figure}[t]
\begin{subfigure}{0.49\textwidth}
    \small
\centering
    \begin{tabular}{lp{.65cm}p{.65cm}p{.65cm}p{.75cm}}
  \toprule
                     & \multicolumn{3}{c}{CoNLL-10} & \multicolumn{1}{c}{JFLEG} \\
    System           & \multicolumn{1}{c}{P} & \multicolumn{1}{c}{R} & \multicolumn{1}{c}{\mtwo} & \multicolumn{1}{c}{GLEU} \\
  \midrule
    Human Avg.       & 73.17      & \bf{68.75} & 72.15 & 62.38 \\
  \midrule
    Ch\&Ng'17        & {79.46}    & 43.73      & 68.29 & 56.78 \\
    \quad Ratio (\%) &  1.08      &  0.64      & 94.66 & 91.02 \\
  \midrule
    This work        & \bf{83.15} & 46.97      & 72.04 & 61.50 \\
    \quad Ratio (\%) &  1.14      &  0.68      & \bf{99.85} & \bf{98.59} \\
  \bottomrule
  \end{tabular}
\end{subfigure}\vspace{.35cm}
\small%

\begin{subfigure}{0.27\textwidth}
\centering%
\begin{tikzpicture}
\begin{axis}[
every axis y label/.style={ at={(ticklabel* cs:0.93)}, anchor=south east, align=right },
height=0.24\textheight,
width=\textwidth,
enlarge y limits,
ymajorgrids, xmajorgrids,
major grid style={dotted},
xtick={1,2,3},
xticklabels={Ch\&Ng'17,This work,Human},
xmin=0.2,
xmax=3.8,
ymin=63,
ymax=78,
xticklabel style={align=right, rotate=45, anchor=north east},
every node near coord/.append style={anchor=south, font=\small, /pgf/number format/.cd, fixed, fixed zerofill, precision=1, /tikz/.cd},
]
\addplot+[black,solid, mark=o, only marks,error bars/.cd,y dir=both,y explicit, error bar style={line width=1pt},
    error mark options={
      rotate=90,
      mark size=6pt,
      line width=1pt
    }]
table[x index=0,
      y expr={(\thisrow{avg}},
      y error plus expr={\thisrow{std}},
      y error minus expr={\thisrow{std}},
      ] {human.test2014};
    \node[] at (axis cs: 2,77.5) {CoNLL-10};
\end{axis}%
\end{tikzpicture}%
\end{subfigure}\hspace{-1cm}%
\begin{subfigure}{0.27\textwidth}%
\centering%
\begin{tikzpicture}%
\begin{axis}[
every axis y label/.style={ at={(ticklabel* cs:0.93)}, anchor=south east, align=right },
cycle list name=black white,
yticklabel pos=right,
height=0.24\textheight,
width=\textwidth,
enlarge y limits,
ymajorgrids, xmajorgrids,
major grid style={dotted},
xtick={1,2,3},
xticklabels={Ch\&Ng'17,This work,Human},
xmin=0.2,
xmax=3.8,
ymin=53,
ymax=68,
legend pos=south east,
legend cell align=left,
legend style={column sep=2pt, font=\scriptsize, draw=none},
xticklabel style={align=right, rotate=45, anchor=north east},
every node near coord/.append style={anchor=south, font=\small, /pgf/number format/.cd, fixed, fixed zerofill, precision=1, /tikz/.cd},
]
\addplot+[black,solid, mark=o, only marks,error bars/.cd,y dir=both,y explicit, error bar style={line width=1pt},
    error mark options={ rotate=90, mark size=6pt, line width=1pt }]
table[x index=0,
      y expr={(\thisrow{avg}},
      y error plus expr={\thisrow{std}},
      y error minus expr={\thisrow{std}},
      ] {human.jflegtest};

    \node[] at (axis cs: 2,67.5) {JFLEG};
\legend{Avg}
\end{axis}%
\end{tikzpicture}%
\end{subfigure}
    \caption{Comparison with human annotators. The figure presents average \mtwo and GLEU scores with standard deviations.
    }
\label{tab:human}
\end{figure}
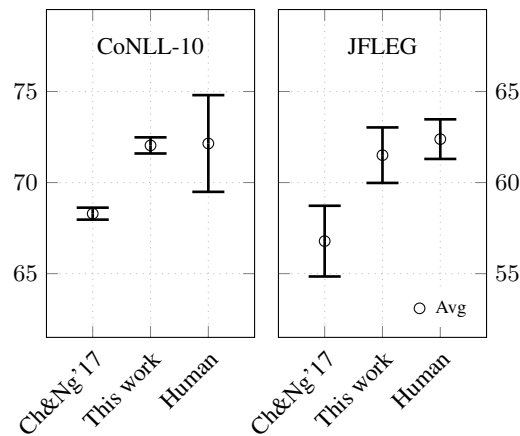

\begin{table*}[!ht]
    \small
  \centering
  \begin{tabular}{ll}
    \toprule
    System  & Example   \\
    \midrule
    Source                 &  \textit{but now every thing is change , the life becom more dificullty .}     \\
    Best SMT               &  \textit{But now \textbf{everything is changed} , the life becom more dificullty .}     \\
    Best NMT               &  \textit{But now \textbf{everything is changing} , the life \textbf{becomes} more \textbf{difficult} .}   \\
    Pipeline               &  \textit{But now \textbf{everything is changed} , the life \textbf{becomes} more \textbf{difficult} .}    \\
    Rescoring              &  \textit{But now \textbf{everything has changed} , the life becom more dificullty .}    \\
    \quad + Pipeline       &  \textit{But now \textbf{everything has changed} , the life \textbf{becomes} more \textbf{difficult} .}   \\
    \midrule
    Reference 1            &  \textit{Now everything has changed , and life becomes more difficult .}       \\
    Reference 2            &  \textit{Everything has changed now and life has become more difficult .}      \\
    Reference 3            &  \textit{But now that everything changes , life becomes more difficult .}      \\
    Reference 4            &  \textit{But now that everything is changing , life becomes more difficult .}  \\
  \bottomrule
  \end{tabular}
  \caption{System outputs for the example source sentence from the JFLEG Test set.}
  \label{tab:examples}
\end{table*}

\section{Analysis and future work}
\label{sec:analysis}

For both benchmarks our systems are close to automatic evaluation results that have been claimed to correspond to human-level performance on the CoNLL-2014 test set and on JFLEG Test. 

\paragraph{Example outputs}
Table~\ref{tab:examples} shows system outputs for an example source sentence from the JFLEG Test corpus that illustrate the complementarity of the statistical and neural approaches.
The SMT and NMT systems produce different corrections.
Rescoring is able to generate a unique correction (\errcor{is change}{has changed}), but it fails in generating some corrections from the neural system, e.g.~misspellings (\err{becom} and \err{dificullty}).
Pipelining, on the other hand, may not improve a local correction made by the SMT system (\err{is changed}).
The combination of the two methods produces output, which is most similar to the references.

\paragraph{Comparison with human annotations}

\newcite{bryant2015how} created an extension of the CoNLL-2014 test set with 10 annotators in total, JFLEG already incorporates corrections from 4 annotators. Human-level results for \mtwo and GLEU were calculated by averaging the scores for each annotator with regard to the remaining 9 (CoNLL) or 3 (JFLEG) annotators, respectively. 

Figure~\ref{tab:human} contains human level scores, our results, and previously best reported results by \newcite{chollampatt2017connecting}. Our best system reaches nearly 100\% of the average human score according to \mtwo and nearly 99\% for GLEU being much closer to that bound than previous works\footnote{During the camera-ready preparation, \newcite{chollampatt2018mlconv} have published a GEC system based on a multilayer convolutional encoder-decoder neural network with a character-based spell-checking module improving the previous best result to 54.79 \mtwo on CoNLL-2014 and 57.47 GLEU on JFLEG Test.}.

Further inspection reveals, however, that the precision/recall trade-off for the automatic system indicates lower coverage compared to human corrections --- lower recall is compensated with high precision\footnote{A similar imbalance between precision and recall is visible on JFLEG when the \mtwo metric is used.}.
Automatic systems might, for example, miss some obvious error corrections and therefore easily be distinguishable from human references. Future work would require a human evaluation effort to draw more conclusions.


\section*{Acknowledgments}

This work was partially funded by Facebook. The views and conclusions contained herein are those of the authors and should not be interpreted as necessarily representing the official policies or endorsements, either expressed or implied, of Facebook.

\bibliography{gecnmt}
\bibliographystyle{acl_natbib}

\end{document}